\documentclass{article}

\PassOptionsToPackage{numbers, compress}{natbib}\usepackage[preprint]{neurips_2019}

\usepackage[utf8]{inputenc} %
\usepackage[T1]{fontenc}    %
\usepackage{hyperref}       %
\usepackage{url}            %
\usepackage{booktabs}       %
\usepackage{amsfonts}       %
\usepackage{nicefrac}       %
\usepackage{microtype}      %

\usepackage{amsmath,amsfonts,bm}

\def\1{\bm{1}}

\def\vp{{\bm{p}}}

\DeclareMathAlphabet{\mathsfit}{\encodingdefault}{\sfdefault}{m}{sl}
\SetMathAlphabet{\mathsfit}{bold}{\encodingdefault}{\sfdefault}{bx}{n}

\DeclareMathOperator*{\argmax}{arg\,max}

\DeclareMathOperator{\sign}{sign}

\usepackage{tabu}
\usepackage{caption}
\usepackage{multirow}
\usepackage{algorithm, algorithmic}
\usepackage{mathtools}
\usepackage{listings}
\usepackage{xspace}
\usepackage{wrapfig}
\usepackage{fontawesome}
\newcommand{\system}{RS-DQN\xspace}
\newcommand{\defend}{\faShield\xspace}
\newcommand{\adversary}{\faUserSecret\xspace}
\newcommand{\clinelight}[1]{\tabucline[lightgrey]{#1}}

\newcommand{\highlight}[2]{\colorbox{#1}{#2}}
\newcommand{\highlightBlock}[3]{\highlight{#1}{\parbox{#2}{#3}} \vspace{-1mm} }

\usepackage[usenames, dvipsnames]{color} %
\usepackage{xcolor} %

\definecolor{my-full-blue}{HTML}{1F77B4}
\definecolor{my-full-orange}{HTML}{FF7F0E}
\definecolor{my-full-green}{HTML}{2CA02C}
\definecolor{my-full-red}{HTML}{d62728}
\definecolor{my-full-purple}{HTML}{9467bd}

\colorlet{my-blue}{my-full-blue!30}
\colorlet{my-orange}{my-full-orange!30}
\colorlet{my-green}{my-full-green!30}
\colorlet{my-red}{my-full-red!30}
\colorlet{my-purple}{my-full-purple!30}
\colorlet{my-gray}{gray!8}

\definecolor{darkred}{rgb}{0.5,0,0}
\definecolor{darkgreen}{rgb}{0,0.5,0}
\definecolor{darkblue}{rgb}{0,0,0.5}
\definecolor{lightgrey}{rgb}{0.7,0.7,0.7}
\definecolor{lightergrey}{rgb}{0.93,0.93,0.93}

\usepackage{acro} %
\DeclareAcronym{cli} {
    short = CLI,
    long = Command Line Interface,
    class = abbrev
}

\usepackage{titlecaps}
\Resetlcwords
\Addlcwords{a after along an and are around as at but by etc}
\Addlcwords{for from if in is nor of on the so to with without yet}
\Addlcwords{}

\newcommand{\eg}{e.g., }
\newcommand{\ie}{i.e., }

\usepackage{lipsum}

\ifdefined\isoverfull
\else
\fi

\usepackage{tikz}
\usetikzlibrary{calc,decorations,decorations.pathmorphing,shapes}
\usetikzlibrary{positioning,shapes,fit,arrows}
\usetikzlibrary{decorations.markings,matrix}
\usetikzlibrary{shapes,shapes.geometric,shapes.callouts,shapes.arrows,shapes.misc,}
\usetikzlibrary{shadows,patterns, shapes.multipart}
\usetikzlibrary{backgrounds,decorations.pathreplacing,automata}

\usetikzlibrary{shapes,arrows, fit, calc, positioning}
\usetikzlibrary{arrows.meta}
\tikzset{>={Latex[width=1.5mm,length=1.2mm]}}

\tikzstyle{l} = [draw, -latex', ->]

\tikzstyle{env} = [rectangle, draw, fill=red!40,
    text width=6em, text centered]
\tikzstyle{algor} = [rectangle, draw, fill=blue!20,
    text width=5em, text centered, rounded corners]
\tikzstyle{net} = [draw, rectangle, rounded corners, fill=yellow!20, node distance=3.5cm]

\tikzstyle{train} = [draw, rectangle, rounded corners, fill=green!10, node distance=3.5cm]

\tikzset{%
    code/.style={fill=gray!8, inner ysep=1pt, inner xsep=15.3pt, draw=gray!8, text width=0.85\linewidth},
    arrow/.style={-{latex[scale=3.0]},out=0,in=180},
    cross/.style={cross out, draw=red, minimum size=2*(#1-\pgflinewidth), inner sep=0pt, outer sep=0pt},
    thickArrow/.style={-, color=gray!10, draw=gray!50, line width=1mm, -triangle 90,postaction={draw, line width=4mm, shorten >=2mm, -}}
}

\usepackage[capitalize]{cleveref}
\crefformat{section}{\S#2#1#3}

\title{Online Robustness Training for\\Deep Reinforcement Learning}

\author{%
  Marc Fischer\\
  Department of Computer Science\\
  ETH Zurich, Switzerland\\
  \texttt{marc.fischer@inf.ethz.ch} \\
  \And
  Matthew Mirman\\
  Department of Computer Science\\
  ETH Zurich, Switzerland\\
  \texttt{matthew.mirman@inf.ethz.ch} \\
  \And 
  Steven Stalder\\
  Department of Computer Science\\
  ETH Zurich, Switzerland\\
  \texttt{staldest@student.ethz.ch} \\
  \And
  Martin Vechev\\
  Department of Computer Science\\
  ETH Zurich, Switzerland\\
  \texttt{martin.vechev@inf.ethz.ch} \\  
}

\begin{document}

\maketitle

\begin{abstract}
In deep reinforcement learning (RL), adversarial attacks can trick an agent into unwanted states and disrupt training. We propose a system called Robust Student-DQN (\system), which permits  online robustness training alongside $Q$ networks, while preserving competitive performance. We show that \system can be combined with (i) state-of-the-art adversarial training and (ii) provably robust training to obtain an agent that is resilient to strong attacks during training and evaluation.

\end{abstract}

\section{Introduction}

To ensure Reinforcement Learning (RL) agents behave reliably in the wild, it is important to consider settings where an adversary aims to interfere with the decisions of the agent.
Most recent progress in RL has focused on handling continuous states via neural networks \cite{sutton2000policy, DQN, peters2008natural}. However, small perturbations to the input of neural networks can yield vastly different outputs \cite{szegedy2013intriguing}. These perturbations are also applicable to neural networks deployed in RL, leading to possible security risks \cite{policyAttacks}.

Existing work in the field of robust RL has focused on small, physically plausible perturbations, discrete states \cite{morimoto2005robust, boyan1995generalization}, or settings with few inputs \cite{mandlekar2017adversarially, gu2018adversary, robustAuto, onlinerobustness, shashua2017deep}. These approaches, however, do not scale to handling gradient-based attacks on large images as studied in supervised classification \cite{szegedy2013intriguing, goodfellow2014explaining, madry2017towards}.

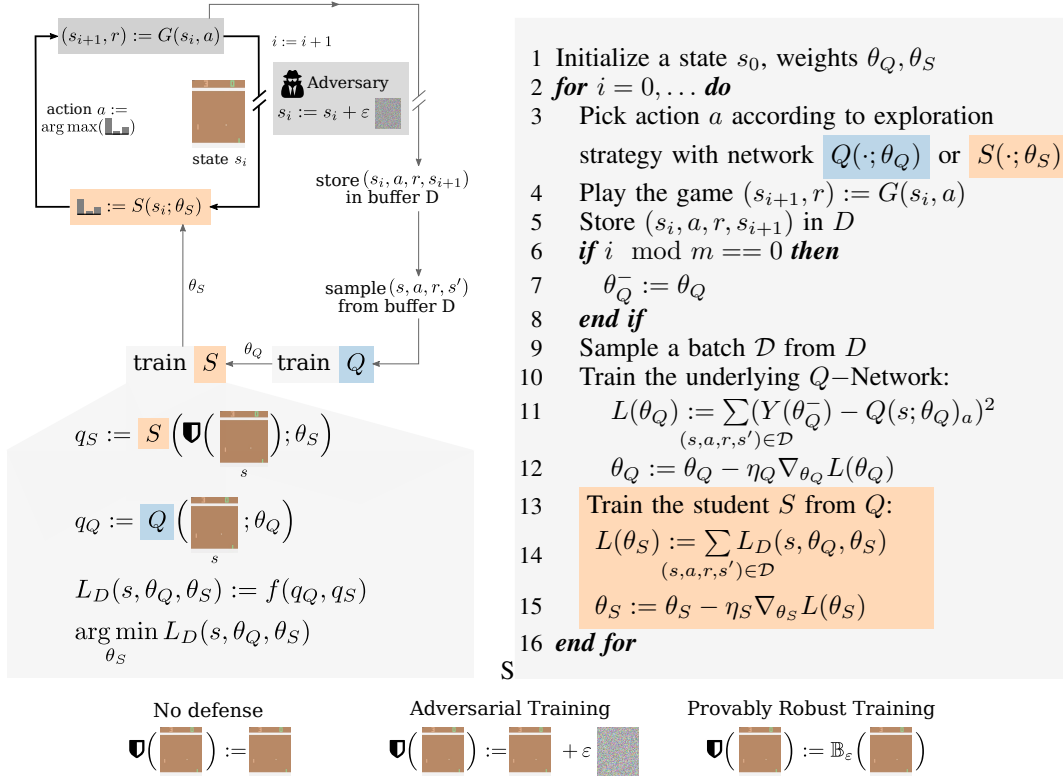
\begin{figure}[!ht]

  \begin{center}
    \begin{minipage}{0.445\linewidth}
      \begin{center}
        \includegraphics[width=\linewidth]{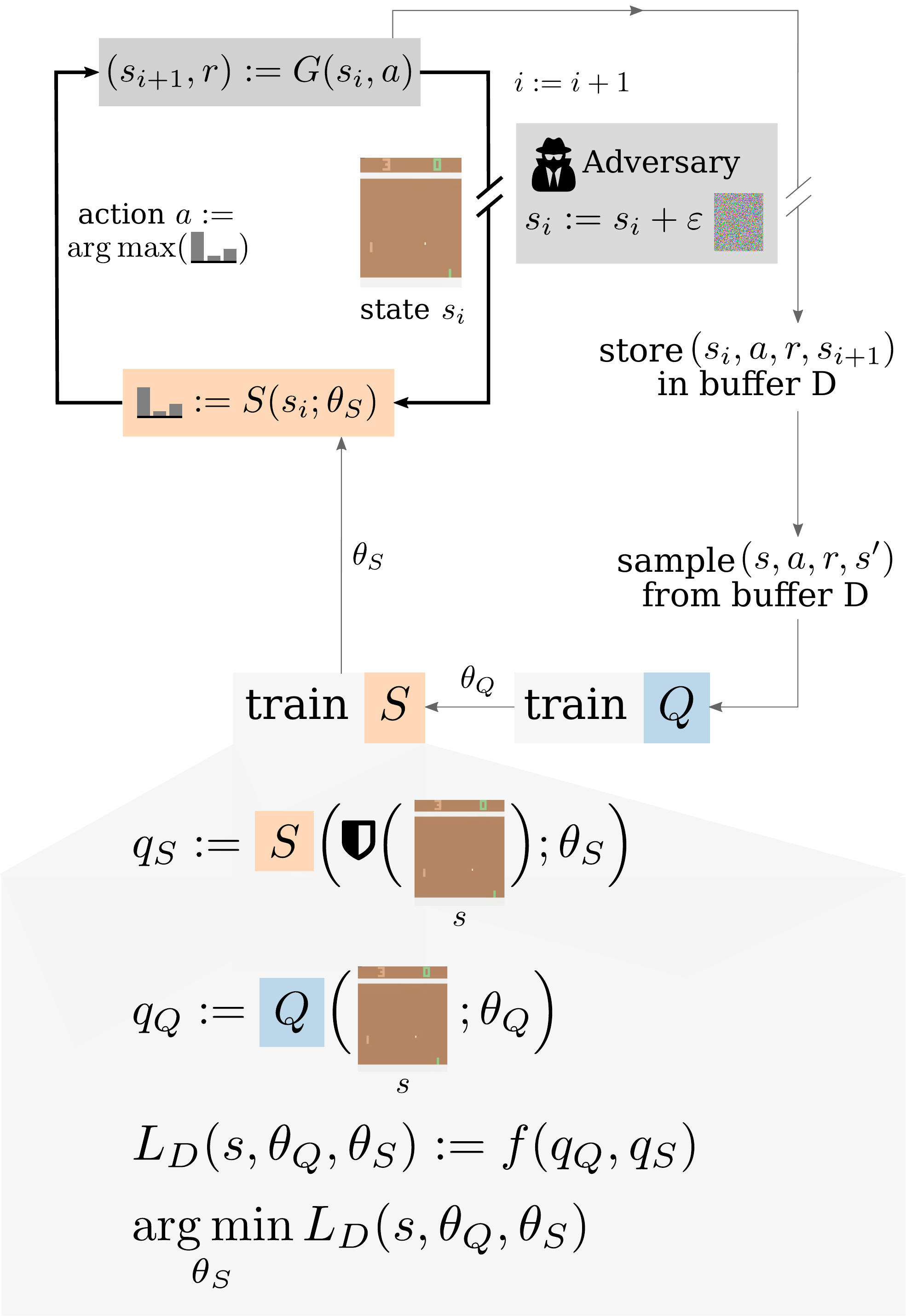}
      \end{center}
    \end{minipage}
    \hfill
    \begin{minipage}{0.53\linewidth}
      \begin{center}
        \vspace{2.3mm}
        S\begin{tikzpicture}
\node[code] (P) {{
\begin{lstlisting}[escapeinside={(*}{*)},
  numberstyle=\small,
  basicstyle=\normalsize,
  columns=fullflexible,
  tabsize=2,
  linewidth=1.115\linewidth,
  numbersep=5pt,
  breaklines=true,
  breakindent=0pt,
  keywordstyle=\color{black}\bfseries\em,
  keywords={, for, if, end if, then, do, end for, end},
  numbers=left
  ]
Initialize a state (*$s_{0}$*), weights (*$\theta_Q,\theta_S$*)
for (*$i=0,\ldots$*) do
	Pick action (*$a$*) according to exploration strategy with network (*\highlight{my-blue}{$Q(\cdot; \theta_Q)$}*) or (*\highlight{my-orange}{$S(\cdot; \theta_S)$}*)
	Play the game (*$(s_{i+1},r) := G(s_{i}, a)$*)
	Store (*$(s_{i},a,r,s_{i+1})$*) in (*$D$*)
	if (*$i \mod m == 0$*) then
		(*$\theta_Q^- := \theta_Q$*)
	end if
	Sample a batch (*$\mathcal{D}$*) from (*$D$*)
	Train the underlying (*$Q$*)-Network:
		(*$\;L(\theta_Q) := \sum\limits_{\mathclap{(s,a,r,s') \in \mathcal{D}}} (Y(\theta^{-}_{Q}) - Q(s;\theta_Q)_a)^2 $*)
		(*$\;\theta_Q := \theta_Q - \eta_Q \nabla_{\theta_Q} L(\theta_Q)$*)
	(*\highlightBlock{my-orange}{4.5cm}{Train the student $S$ from $Q$:}*)
	(*\highlightBlock{my-orange}{4.5cm}{	$\;L(\theta_S) := \sum\limits_{\mathclap{(s,a,r,s') \in \mathcal{D}}} L_{D}(s,\theta_Q,\theta_S)$}*)
	(*\highlightBlock{my-orange}{4.5cm}{	$\;\theta_S := \theta_S - \eta_S \nabla_{\theta_S} L(\theta_S)$} \vspace{1mm}*)
end for
\end{lstlisting}}};
\end{tikzpicture} %
      \end{center}
\end{minipage}\\[0.7em]
\includegraphics[height=1cm]{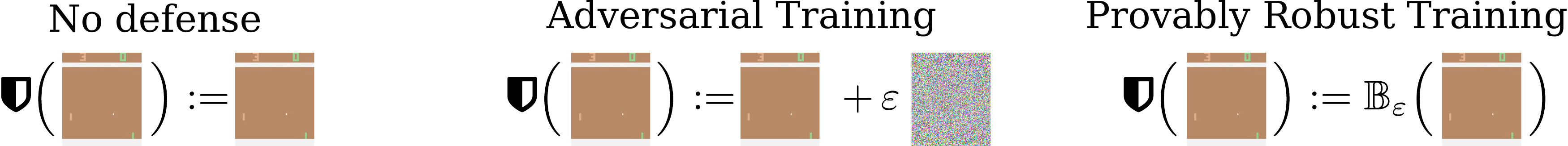}
\end{center}

\caption{(left) Overview of of \system. (right) Simplified pseudo-code for DQN and Student-DQN training. Parts in blue are specific to \highlight{my-blue}{DQN} and parts in orange to \highlight{my-orange}{Student-DQN}.} \label{fig:alg}

\end{figure}

In this work we present a new approach for training RL systems to be more robust against adversarial perturbations.
The key idea, shown in \cref{fig:alg}, is to split the standard DQN architecture into a policy (student) network $S$ and a $Q$ network in a way which enables us to robustly train the policy network $S$ and use it for exploration, while at the same time preserving the standard way of training the $Q$ network.
We then show how to naturally incorporate state-of-the-art defenses developed in supervised deep learning to the setting of reinforcement learning, by training the student network in two ways: (i) via adversarial training with methods such as PGD \cite{madry2017towards} where we generate adversarial states that decrease the chance the optimal action is selected and use them to train the policy network, and (ii) via provably robust training with symbolic methods \cite{diffai, diffaiNew} which guarantee the network will select the right action in a given state despite any possible perturbation (within a range) of that state.

\paragraph{Key Contributions} Our main contributions are:
\begin{itemize}
\item A novel deep RL algorithm, \system, designed to be defended with state-of-the-art adversarial training as well as provably robust training.
\item We show that when no attack is present, \system and DQN obtain similar scores, while in the presence of attacks, undefended DQNs fail while \system remains robust.
\item An evaluation which demonstrates that \system can produce an agent that is certifiably robust to $\pm 1$ pixel intensity changes on Atari games with scores comparable to DQN.
\end{itemize} 

\section{Overview of the \system Architecture} \label{sec:background}
We now provide an overview of \system, shown in \cref{fig:alg}, and introduce its key ingredients.
The top left part of \cref{fig:alg} shows an agent $S$ playing a game $G$ by observing the current state $s_{i}$ and picking an action $a$. Based on the state and the action, the game transitions to the next state $s_{i+1}$ with a reward $r$. These interactions $(s_{i}, a, r, s_{i+1})$ are stored and later used in training. If we were to replace the agent $S$ (discussed next) with the network $Q$, we would end up with the standard loop used in DQNs.

A key difference to standard training is the presence of an adversary (\adversary) which introduces perturbations to state $s_{i}$. This can make the agent select sub-optimal actions leading to lower rewards.

\cref{fig:alg} also outlines the training of both networks, $S$ and $Q$, via the pseudo-code on the right (discussed shortly), also pictorially illustrated below the play-loop in the gray area. Here, the network $Q$ is trained in the standard DQN manner, however \system also trains an additional (student) network $S$ by distilling $Q$'s policy. We note that this distillation step is not a defense by itself, however it does allow us to incorporate state-of-the-art defensive training measures (\defend) to $S$, leading to an agent $S$ that can robustly interact with the game $G$ and a potential adversary. At a high level, these defenses (shown at the bottom of the picture) take as input the current state and produce a new state, either by adding noise (adversarial training) or by defining a symbolic region that captures a set of perturbations (provably robust training).
 The split of $S$ and $Q$ is critical for the application of defenses, as adversarial training directly on $Q$ severely hurts its performance  \cite{whatDoesntKillyou, robustDeep} and the application of existing provably robust training methods to DQN training is not straightforward.
We discuss defenses in detail later in the paper.

\subsection{Deep Reinforcement Learning}
We now briefly introduce standard deep Q-Learning and policy distillation.
Formally, the goal of reinforcement learning is to determine a policy for a given game $G$.
The policy represents an agent and determines which actions it selects. The function $Q^{*}(s, a)$ describes the expected (discounted) future reward that an ideal agent can obtain, when it selects an action $a$ in state $s$. The objective of Q-Learning \cite{watkins1992q} is to approximate $Q^*$ and construct a policy where the agent greedily selects the action with the maximal value.

\paragraph{Deep Q-Learning}
In Deep Q-Learning \cite{mnihHumanlevelControlDeep2015, DQN}, $Q^*$ is progressively approximated with weights $\theta_{Q}$ for a neural network $Q(s, a;\theta_{Q})$ --- referred to as a deep Q network (DQN). Because $Q(s; \theta_{Q})$ produces a vector of scores for all actions, we write $Q(s;\theta_{Q})_a$ instead of $Q(s, a;\theta_{Q})$.

The pseudo-code in \cref{fig:alg}, when the orange boxes are ignored, shows the standard algorithm for training DQNs.
As discussed before, the agent $Q$ interacts with the game $G$ and the resulting state transition is stored in the experience replay buffer $D$ (lines~3-5). During training, this interaction involves an exploration strategy \eg \emph{$\epsilon$-greedy} where the action $\argmax_a Q(s_{i}; \theta_{Q})_a$ is chosen with probability $1 - \epsilon$ and a random action with probability $\epsilon$ (for $\epsilon \in [0, 1]$).
Next, on lines~9-12, the weights $\theta_{Q}$ of the DQN are updated via stochastic gradient decent on $L(\theta_{Q})$ over a batch $\mathcal{D}$ sampled from $D$.
Integral to the DQN algorithm is the use of a network with lagging weights $\theta_{Q, i}^- = \theta_{Q, m \lfloor \frac{i}{m} \rfloor}$ for some $m$, which is used to calculate the target score $Y(\theta_{Q}^{-}) = r + \gamma \max_{a'} Q(s';\theta_{Q}^{-})_{a'}$.

When applying DQN training to video games, one iteration of the for-loop is referred to as a frame.
Playing one game to completion (after which it is reset) is an episode.

\paragraph{Policy Distillation for DQN training}
It is possible to improve the learned policy in Deep Q-Learning using {\em{Policy Distillation}}\cite{rusuPolicyDistillation2015}. In this method, a $Q$-approximation is first learned by the standard DQN algorithm. New games are then played using $Q$ as a greedy policy and the states $s$ are recorded. A student network $S$ is then trained \emph{offline} on these states so to mimic the behavior of $Q$. The main application of this method is to train a much smaller network $S$ while retaining the performance of a previously trained $Q$ network.

In this work we introduce a new method for combining DQN \emph{training} with Policy Distillation. Unlike \cite{rusuPolicyDistillation2015}, with our method: (i) the learning of the student is performed online, (ii) the student is actively involved in training as it affects the replay buffer, and (iii) both $S$ and $Q$ use the same architecture.

Concretely, lines~13-15 of \cref{fig:alg} show how to apply policy distillation to train the student network $S$ from $Q$, using the distillation loss $L_{D}$ (the exact loss is discussed in \cref{sec:student}). The process is also pictorially illustrated on the left part of the figure. Note that in our setting, we do not require $S$ to be smaller than $Q$, we simply need $S$ to be able to apply defensive training methods to it.

Importantly, we remark that while we build on top of policy distillation to produce a robust neural network, our method is distinct from \emph{defensive distillation} \cite{defensiveDistillation}, which is known to be ineffective in producing robust neural networks \cite{carlini2016defensive}. We use policy distillation only as a first step to enabling strong defenses (\defend).

In related work, distillation has also been used device an collaborative RL algorithm \cite{linCollaborativeDeepReinforcement2017},  allowing for knowledge transfer between multiple agents playing simultaneously in different environments with potentially different tasks. Similar to distillation, a DQN algorithm \cite{chenAgentAwareDropoutDQN2017} where the agent predicts when to consult and how to learn from a pre-defined rule-based teacher policy have been proposed.

\subsection{Adversarial Attacks \& Defenses} \label{sec:background_attack}

It is known that neural networks are susceptible to adversarial perturbations\cite{szegedy2013intriguing}: inputs similar to genuine ones that lead to different neural network outputs. A common method to compute such perturbations is the Fast Gradient Sign Method (FGSM) \cite{goodfellow2014explaining}, which finds an adversarial input $x'$ s.t. 
\[x' \in \mathbb{B}_\varepsilon(x) = \{ x'\; | \; ||x - x'||_\infty \leq \varepsilon \}.\]
Here, $\mathbb{B}_\varepsilon(x)$ is an $\varepsilon$-sized $L_\infty$ ball around $x$ (not to be confused with the $\epsilon$ used for $\epsilon-$greedy exploration). For a network $N$, input $x$, and label $t$, untargeted \text{FGSM} is defined as:
\begin{align}
x' = \text{FGSM}_\varepsilon(x, t, N) = x + \varepsilon \cdot \sign(\nabla_x \mathcal{H}(\sigma(N(x), t) ) \label{eq:fgsm}
\end{align}

where $\mathcal{H}(\vp, t)$ denotes the standard cross-entropy loss between two probability distributions as used in classification and $t$ a onehot-encoded distribution. We let \(\sigma(\cdot)\) denote the softmax function, assuming the outputs of network \(N\) to be logits or equivalent scores such as Q-values.

This version of FGSM produces $x' \in \mathbb{B}_\varepsilon(x)$ where $x'$ has a high chance of not being classified to label $t$.
Since \(t\) is the correct label for \(x\), a successful attack will lead to \(N\) treating \(x\) as \emph{anything but} \(t\) --- an \emph{untargeted} attack. Attacks that lead to \(N\) treating \(x'\) as a specific \(t'\) are called \emph{targeted}.

A stronger version of this attack is called Projected Gradient Decent (PGD) \cite{madry2017towards}.
\(\text{PGD}_{\varepsilon}(x, t, N, k)\) denotes an attack where \(\text{FGSM}_{\frac{\varepsilon}{k}}\) is applied \(k\) times successively (with an additional projection step).
We denote it as \(\text{PGD}(x, t, N)\) in the paper and specify the value for \(\varepsilon\) and \(k\) when needed.

\paragraph{Adversarial Attacks and Defenses in RL}
In the case of RL, an adversary (\adversary in \cref{fig:alg}) can apply attacks both, at training and testing time. At testing time, untargeted attacks can lower the reward attained by an RL agent while targeted attacks can be used to guide it into specific states \cite{advTactics}. Attacks during training can significantly lower the reward the agent is capable of attaining and even prevent learning altogether \cite{whatDoesntKillyou} --- especially in games with high-dimensional inputs such as the frames of Atari games. Further, this effect can be intentionally used to prevent learning \cite{behzadan2017vulnerability}.

Adversarial training (AT) \cite{madry2017towards} --- which aims to make a neural network robust to adversarial perturbations --- requires to deliberately attack the network during training. While this yields more robust networks, in Deep Q-Learning it can degrade the performance of the agent to the point where it fails in learning to play the game. In \cref{sec:defenses}, we show the incorporation of AT as an instantiation of \defend in \cref{fig:alg}.
A version of AT has previously been applied to DQNs, improving its experimental robustness \cite{robustDeep} for low-dimensional inputs, but incurred large reward drops from attacks.
Recently, \cite{CertifiedAdversarialRobustness} leveraged techniques from the robustness analysis of neural networks to  choose more robust actions from a trained DQN.

\section{Training Student-DQN} \label{sec:student}

We now proceed to formally introduce our Student-DQN architecture. As discussed in \cref{sec:defenses} later, this architecture enables the incorporation of constraints into the training process, for example state-of-the-art adversarial defenses \cite{madry2017towards, diffai}. The algorithm consists of: a standard DQN $Q$, a student network $S$, a loss $L_D$, and learning rates $\eta_Q$, $\eta_S$. As with standard DQN training, Student-DQN alternates between playing the game and training.

The pseudo-code in \cref{fig:alg} shows the training of Student-DQN and highlights the differences to standard DQN training: we use one network for exploration and another for learning the Q-function. In line~3 in \cref{fig:alg}, the exploration is first performed by the student network $S$, then the $Q$ network is trained in the standard way (line~9-12) with the learning rate $\eta_{Q}$, and finally (lines~13-15) the $S$ network is distilled from $Q$ using the distillation loss $L_D$ and learning rate $\eta_{S}$. In contrast, in standard DQN training, we would not use the $S$ network but would explore directly using the $Q$ network and would again train the $Q$ network in the same way (line~9-12).

As both $S$ and $Q$ are expected to learn the underlying Q-function, both networks could be used at testing time. However, since we will be training $S$ with additional constraints, we will deploy $S$.

The intuition behind this algorithm is that the loss for the $Q$ network has not been changed and thus after being trained on sufficiently many random samples, the $Q$ function will approach $Q^*$ regardless of what the student learns. Theoretical results show that Q-Learning algorithms learn the correct Q-function given a sufficiently random exploration agent \citep{even2002convergence, bertsekas2008neuro, tsitsiklis1994asynchronous, watkins1992q}.

Assuming the student is capable of learning \(Q^{*}\),
it should be able to handle the concept-shift of $Q$ learning and incorporate the
defense.

Ideally, the additional constraints will allow the student to achieve a better score when playing, so that it explores higher reward paths.

Compared to standard DQN training, the Student-DQN algorithm stores one additional network and performs two neural network updates in a training step instead of one. Thus, the asymptotic complexity of both training methods is the same. In \cref{sec:evaluation} we empirically show that both systems run at similar speeds in practice and exhibit comparable sample complexity.

\paragraph{Loss}

Student-DQN admits many possible choices for the distillation loss $L_D$ such as the mean squared error (MSE) loss and the cross-entropy (CE) loss. That is, we can instantiate $L_D$ as:
\begin{align}
&L_{\text{MSE}}(s, \theta_Q, \theta_S) = ||Q(s; \theta_Q) - S(s; \theta_S)||^2_2 \nonumber\\
  &L_{\text{CE}}(s, \theta_Q, \theta_S) = \mathcal{H} \big( \sigma( S(s; \theta_S) ), \argmax_{a \in \mathcal{A}} Q(s; \theta_Q)_{a} \big)  \label{eq:ce}
\end{align}

For \(L_{\text{MSE}}\), the outputs of the \(S\) network are treated as Q-values, while in \(L_{\text{CE}}\), the  output of \(S\) is treated as the logits of a probability distribution.
Both losses, along with a third based on KL-divergence, were described in the context of policy distillation (\(L_{\text{CE}}\) is the same as \(L_{\text{NLL}}\) \cite{rusuPolicyDistillation2015} up to scale). Note that only \(L_{\text{MSE}}\) will produce a student \(S\) that gives the same numerical result as \(Q\). However, for the other losses, \(S\) will still learn to take the same decisions as \(Q\). In previous work on policy distillation \cite{rusuPolicyDistillation2015}, the KL-based loss worked best, however we found the CE-based loss to be most suitable for \system and we only discuss the adaption of $L_{\text{CE}}$ going forward.

We note that many of the extensions to the standard DQN algorithm can also be applied to \system unchanged \cite{prioreplay,doubledqn}. However, extensions such as \emph{DuelingDQN}\cite{duelingdqn} and \emph{NoisyNet} \cite{fortunatoNoisyNetworksExploration2017} do require small adaptations. We find that both of these extensions help training and further, the \emph{DuelingDQN} method is particularly critical when incorporating provably robust training.

\paragraph{Incorporating \emph{DuelingDQN}}
\emph{DuelingDQN} \cite{duelingdqn} introduces a specific $Q$ network architecture, which has been observed to achieve better learning performance. Specifically, \emph{DuelingDQN} consisting of two components: an advantage network $A(s;\theta_{Q}) \colon \mathbb{R}^{|\mathcal{A}|}$ computing the relative advantage of the $|\mathcal{A}|$ actions, and a value network $V(s;\theta) \colon \mathbb{R}$.  These are combined to obtain

\begin{equation}
Q(s;\theta_{Q})_a = V(s ;\theta_{Q}) + (A(s;\theta_{Q})_a - \frac{1}{|\mathcal{A}|} \sum_{a' \in \mathcal{A}} A(s;\theta_{Q})_{a'}) \label{eq:duelingQ}
\end{equation}

where $\mathcal{A}$ denotes the action space. The advantage and value networks are defined to share the early layers. We note that the output of $A$ alone is sufficient to replicate a greedy policy based on $Q$. However, the value network may be able to better discriminate between states and thus aid in training.

To incorporate \emph{DuelingDQN}, both $Q$ and $S$ use the DuelingDQN-architecture described by \cref{eq:duelingQ}. To instantiate $L_D$ in \cref{fig:alg} accordingly, we define

\begin{equation}
    L_{\text{CE}}^{\text{Duel}}(s, \theta_Q, \theta_S) = \mathcal{H} \big( \sigma(A(s; \theta_S)), \argmax_{a} A(s; \theta_Q)_{a} \big) + || V(s;\theta_S) - V(s;\theta_Q) ||_2^2 \label{eq:ce_duel}
\end{equation}

as an adaption of $L_{\text{CE}}$ (\cref{eq:ce}).
In practice we also found a hybrid version of \cref{eq:ce,eq:ce_duel} useful, which uses the cross-entropy term as in \cref{eq:ce} plus the loss on the value-networks as in \cref{eq:ce_duel}.

\paragraph{Incorporating NoisyNet}
While \emph{$\epsilon$-greedy} exploration comes with theoretical guarantees, \emph{NoisyNet} \cite{fortunatoNoisyNetworksExploration2017}, an extension to DQNs where the $Q$ network learns mean and variance of Gaussian noise on the weights in its dense layers, was prosposed. Using samples from the noise distributions as the source of randomness during exploration, this exploration strategy allows agents to achieve higher scores.

In Student-DQN, only the student network $S$ utilizes noise while the $Q$ network uses non-noisy weights. We notice that since $S$ lags behind $Q$, it might require more diverse samples even when $Q$ has learned the correct behavior sufficiently well. We thus introduce an exploration noise constant $\kappa \geq 1$, which during exploration is multiplied with the weight variances.

\section{Training Robust Student-DQN (\system)} \label{sec:defenses}

The primary motivation for decoupling the DQN agent into a policy-student network $S$ and a $Q$ network is to enable one to leverage additional constraints on $S$ without strongly affecting learning of the correct Q-function.

Concretely, we address the problem of adversarial robustness in reinforcement learning by improving the robustness of $S$. Specifically, we do this by instantiating $L_{D}$ in \cref{fig:alg} with different defenses (\defend): adversarial training and provably robust training.

\paragraph{Incorporating Adversarial Training}

A variety of techniques have been developed for increasing the robustness of neural networks, typically by training with adversarial examples \citep{tramer2017ensemble,shaham2015understanding, madry2017towards}.
An attack is utilized to deliberately produce adversarial examples, which are then used in training. Rather than providing the $Q$ network with adversarial examples as in \cite{mandlekar2017adversarially}, we only provide them to $S$.
Formally we do this by adapting the loss $L_{\text{CE}}$ (\cref{eq:ce}):

\begin{equation}
L_{\text{CE, def}}(s, \theta_Q, \theta_S) = \mathcal{H} \big( \sigma (S(s_{d}; \theta_S)), \argmax_{a} Q(s; \theta_Q)_{a} \big)  \label{eq:ce_adv}
\end{equation}

Here $s_d = \text{PGD}(s, \argmax_a S(s;\theta_S)_a, S(\cdot;\theta_S))$. While the computation of $s_d$ could be inlined into \cref{eq:ce_adv} we refrain from this here so to clarify that the gradient is not propagated through the PGD attack and $s_d$ is treated as a constant.
For DuelingDQNs we calculate the advantage and value separately, adapting \cref{eq:ce_duel}:

\begin{equation}
    L_{\text{CE, def}}^{\text{Duel}}(s, \theta_Q, \theta_S) = \mathcal{H} \big( \sigma(A(s_{d}; \theta_S)), \argmax_{a} A(s; \theta_Q)_{a} \big) + || V(s_{d};\theta_S) - V(s;\theta_Q) ||_2^2 \label{eq:ce_duel_def}
\end{equation}

We note that while we explicitly use PGD here, our approach is not limited to the kind of attack that is used to provide the adapted training samples.

\paragraph{Incorporating Provably Robustness Training}
While adversarial training produces empirically robust networks, they are usually not provably robust, that is, it cannot be shown that the network takes the same decision on all perturbed inputs that are reasonably close to the original input. Recently, techniques to train networks to be certifiably robust have been introduced. Although initially limited to small networks, multiple techniques have been developed since to train increasingly larger networks to be provably robust \cite{raghunathan2018certified, kolter2018provable, diffai, dvijotham2018training, wong2018scaling, diffaiNew}. These techniques represent different points in the spectrum of accuracy, training speed, certifiable robustness and experimental robustness.

Here we will show how to apply one of these methods \cite{diffai} (DiffAI) as a defense (\defend) for \system. Using DiffAI's Interval abstraction will allow us to effectively train on $\mathbb{B}_\varepsilon(s)$ rather than $s$.
The DiffAI framework has been shown effective in training networks on the scale of DQNs used in Atari games with minimal speed and memory overheads over standard, undefended training.

We use DiffAI to soundly propagate $\mathbb{B}_\varepsilon(s)$ through the network $S(\cdot; \theta_S)$ (via symbolic computation), obtaining a symbolic element $g$ as a result.

Because we are propagating $\mathbb{B}_\varepsilon(s)$ symbolically, we can verify that for a target \(t\), we have that $\forall \bar{s} \in \mathbb{B}_\varepsilon(s) . \argmax_a S(\bar{s};\theta_S)_a = t$. That is, in this case, we can \emph{certify} that for any element inside \(\mathbb{B}_\varepsilon(s)\) the agent \(S\) will pick the same action $t$.

Further, DiffAI defines a differentiable loss $L_I \colon \text{Interval} \times \mathbb{N} \rightarrow \mathbb{R}$ which takes as input the final element $g$ and a target (which is an action in our case) and allows for training the network on  $\mathbb{B}_\varepsilon(s)$.

Using this approach for training requires further modification of our loss function.
We apply DiffAI only to the DuelingDQN version of \system and thus extend \(L_{\text{CE}}^{\text{Duel}}\)-loss (\cref{eq:ce_duel}):
\begin{equation}
  L_{\text{Provable}}(s, \theta_Q, \theta_S) = || V(s;\theta_Q) - V(s;\theta_S) ||_2^2 + \lambda_{D} L_{\text{DiffAI}}(g_{A}(s), \argmax_a A(s; \theta_Q)_a) \label{eq:diffai}
\end{equation}

Here, \(g_{A}(s)\) denotes the symbolic propagation of $\mathbb{B}_\varepsilon(s)$ through $A(\cdot; \theta_S)$ and we have the loss \(L_{\text{DiffAI}}(g_{A}(s), t) = \lambda \mathcal{H} \big( A(s; \theta_S), t \big) + (1-\lambda) L_{I}(g_{A}(s), t)  \) which defines a linear combination of the standard cross-entropy loss and the Interval loss $L_{I}$ for $\lambda \in [0, 1]$. Throughout the learning process, we linearly anneal the value \(\lambda\) to shift more weight on provability.

We note that training the advantage network \(A(\cdot;\theta_{S})\) to be robust is sufficient, as \(V(\cdot;\theta_{S})\) can be disregarded for making decisions and is only important for training.
\begin{table*}[t]
\small
\centering
\caption{Average evaluation scores over 15 games with and without training and test time attacks. \system is defended with Adversarial Training.}
\vspace{5pt}
\begin{tabu}{@{}ll@{\hspace{1cm}}rr@{\hspace{1cm}}rr@{}}

  & & \multicolumn{2}{c@{\hspace{0.5cm}}}{no training attack}
  & \multicolumn{2}{c@{}}{\emph{TrainingPGD(k=1)}}\\
\cmidrule(l{0pt}r{0.5cm}){3-4}
\cmidrule(l{0pt}r{0pt}){5-6}

Game
& TestAttack
& DQN
& \system
& DQN
& \system\\

\midrule

\multirow{3}{*}{Freeway} & none    & 33.00 &  29.33 & 21.73 & 32.93\\
                         & \emph{TestPGD(k=4)} &  0.00 & 27.93 & 22.53 & 32.53\\
                         \clinelight{1-}
\multirow{3}{*}{Bank Heist} & none    & 222.00 &  112.66 & 220.00 & 238.66\\
                         & \emph{TestPGD(k=4)} &  3.33 & 121.33 & 45.33 & 190.67\\
                         \clinelight{1-}
\multirow{3}{*}{Pong} & none    &  20.20 & 16.73  &  20.46 & 19.73\\
                         & \emph{TestPGD(k=4)} &  -20.73 & 15.87 & -12.87 & 18.13\\
                         \clinelight{1-}
\multirow{3}{*}{Boxing} & none    &  95.87 & 93.27 & 84.80 & 80.67\\
                         & \emph{TestPGD(k=4)} &  -2.80 & 70.33 & 9.20 & 50.87\\
                         \clinelight{1-}
\multirow{3}{*}{Road Runner} & none    &  9406.67 & 11920.00 & 7066.67 & 12106.67\\
                         & \emph{TestPGD(k=4)} &  0.00 & 9293.33  & 1266.67 & 5753.33\\
                       \end{tabu}
\label{tab:score}
\end{table*}

\section{Experimental Evaluation} \label{sec:evaluation}
We now present our detailed evaluation of \system, in which we demonstrate that: (i) \system instantiated with either adversarial training or provably robust training is capable of obtaining scores similar to undefended DQN, (ii) \system instantiated with adversarial training is empirically robust to PGD attacks in both training and evaluation, and (iii) \system can be trained to be provably robust to $\pm 1$ pixel intensity changes on Atari game frames.

\subsection{Experimental Setup} \label{sec:ex_setup}
We tested our algorithm with 5 Atari games \cite{bellemare2013arcade} from the OpenAI Gym \cite{brockman2016openai}.

Various combinations of extensions to the DQN algorithm -- known to increase the performance of the agent -- are discussed in \cite{rainbowDQN}. We implemented a subset of these for both DQN and \system: Priority Replay \cite{prioreplay}, DoubleDQN \cite{doubledqn}, DuelingDQN \cite{duelingdqn}, and NoisyNet \cite{fortunatoNoisyNetworksExploration2017}. The extensions of these algorithms to \system were already discussed in Section \cref{sec:student}.

We trained each agent for \(4\) million frames. All further parameters are provided in \cref{tab:hyperparameters} in \cref{sec:hyper}.

We implemented both \system and DQN in PyTorch \cite{paszke2017automatic} asynchronously \cite{mnih2016asynchronous}. On a machine with a Nvidia~1080Ti, our implementations of DQN and Student-DQN, both play with a peek speed of around 266 frames per second without additional attacks or defenses. A training run spanning 4M frames takes between 4 to 30 hours depending on the exact parameters.

\paragraph{Attacks}

In our evaluation we use \emph{TrainingPGD(k=1}) (attack during training) and  \emph{TestPGD(k=4)} (attack during testing) as our attack schemes. \emph{TestPGD} refers to the standard PGD attack as introduced in  \cref{sec:background_attack}. \emph{TrainingPGD} refers to a similar attack where we flip the sign in \cref{eq:fgsm} to be a $-$ rather than a $+$. The intuition behind this attack is that the action of the agent is reinforced rather than changed, creating an illusion of successful training. However, when evaluated without this perturbation (\ie in the testing phase), the performance of the agent will deteriorate.
For all attacks we use \(k\) as indicated in the name and \(\varepsilon = 0.004\), which roughly corresponds to changing a pixel by an 8-bit value ($\frac{1}{255}$).

DQNs use an optimization where the input to the network $s$ is not only the current frame but rather a stack of the last four seen frames in grayscale (see \cite{DQN} for details). We apply our attacks and defenses to these 4-stacks of consecutive frames (of size $84 \times 84$) used for training DQNs on Atari. While single frame attacks conceptually fit the setting, we are primarily evaluating the effective of defenses and thus choose the the more efficient 4-stack-attacks.
In this evaluation we also always compute the attack perturbation based on the exploring agent, which is \(Q\) for DQN and \(S\) for \system.

\subsection{Empirically Robust \system} \label{sec:eval_def}
We now want to show the utility of defending with \system against attacks during training and testing. Thus, we consider two agents, standard DQN and \system (which uses adversarial training).

For \system, we instantiate \(L_{D}\) with $L_{\text{CE, def}}^{\text{Duel}}$ (utilizing the hybrid version of $ L_{\text{CE}}^{\text{Duel}}$ described at the end of \cref{sec:student}), where we defend with Adversarial Training using PGD with $k=1$ and $\varepsilon = 0.004$.

During the training process of each agent, we played a validation episode every 10 episodes. In these validation episodes, we disable noise due to NoisyNet and use $\epsilon$-greedy exploration with $\epsilon = 0.005$. We save the parameters that produce the agent with the highest score in an validation episode.

Afterwards, we evaluate DQN and \system by running another 15 evaluation games with the best performing weight. The resulting average scores can be found in \cref{tab:score}. The columns indicate which algorithm is used and whether it was attacked during training, while the rows show different attacks (none, \emph{TestPGD(k=4)}) during testing. In \cref{sec:results_details}, we report additional numbers for \emph{TestPGD(k=1)} and \emph{TestPGD(k=50)}, as well as the standard deviation of the scores.

The two  columns titled ``no training attack'' in \cref{tab:score} show that without a training attack, \system achieves scores that are only slightly lower than DQN when no attack is applied at test time. However, if \emph{TestPGD(k=4)} is applied, DQN's scores drop significantly while \system's are robust.

In the presence of attacks during training (\emph{TrainingPGD(k=1)}), we observe that DQN and \system attain similar scores without a testing attack. With DQN, we see the effect of the training attack as the attained scores are much lower than without a training attack, while \system behaves similarly in the two scenarios. When evaluated in the presence of \emph{TestPGD(k=4)}, we again see \system remaining robust while DQN scores drop.

\begin{table*}[t]
\small
\centering
\caption{Score and size the largest provable region $\mathbb{B}_{\varepsilon}(s)$ around the state $s$. Both numbers are averaged over 15 games. Values for $\varepsilon$ are found via binary search and multiplied by 255 to correspond to discrete pixel intensity changes. \system uses provably robust training.}

  \vspace{5pt}
\begin{tabu}{@{}lrr@{\hspace{1.5cm}}rr@{}}

  & \multicolumn{2}{c@{\hspace{1cm}}}{}
  & \multicolumn{2}{c@{}}{Size $\varepsilon_{\text{max}}$ of the largest }\\  
  & \multicolumn{2}{c@{\hspace{1cm}}}{Score}
  & \multicolumn{2}{c@{}}{provable interval $\mathbb{B}_{\varepsilon_{\text{max}}}(s)$ }\\
\cmidrule(lr{1.3cm}){2-3}
\cmidrule(l{0pt}r{0pt}){4-5}

Game
& DQN
& \system
& DQN
& \system\\

\midrule

Freeway     & 33.00    &  32.53  & 0.000 & 2.028\\
Bank Heist  & 222.00   & 154.00  & 0.001 & 1.373\\
Pong        &  20.20   &   5.13  & 0.000 & 1.075\\
Boxing      &  95.87   & 90.47   & 0.000 & 1.383\\
Road Runner &  9406.67 & 5166.67 & 0.000 & 1.233\\

\end{tabu}
\label{tab:diffai}
\end{table*}

\subsection{Provably Robust \system}
As introduced in \cref{sec:defenses}, we can use DiffAI to prove robustness for a region of inputs, as well as train a network to be more provable. We will now evaluate a \system agent, trained using \(L_{\text{Provable}}\) (\cref{eq:diffai}).

A recent version of DiffAI \cite{diffaiNew} introduced a language for specifying detailed training procedures. To enable reproducibility, we provide the full parameters of the training procedure expressed in this language in \cref{sec:diffaicommand}. Specifically, we start the learning process with $\lambda = 1$ (all weight on the normal cross entropy loss) and linearly anneal it to be $\lambda = 0$ (all weight on the robustness training) from frame 500000 to frame 4M. Similarly, we also anneal the size of the training region $\varepsilon$ from $0$ to $\frac{1}{255}$ over the same time frame. Further we use $\lambda_{D} = 1$.

The agents were trained and evaluated as explained before. However, since we are increasing the importance of the robustness loss with the frame number, we now use the weights after all of the training procedure has completed. We play evaluation games, as before, but this time also measure the size \(\varepsilon\) of the largest ball \(\mathbb{B}_{\varepsilon}(s)\), for which the action of the agent is robust within the ball --- called $\varepsilon_{\text{max}}$. The values were found by binary search over $\varepsilon$ between \(0\) and \(1\) with up to \(20\) iterations.%

\cref{tab:diffai} reports the score and \(\varepsilon_{\text{max}}\) averaged over 15 evaluation games. The DQN weights were taken from the same highest scoring snapshot as in \cref{sec:eval_def}.
We first observe that the \system agent, trained with DiffAI, obtains similar but slightly lower scores than DQN. This is to be expected as it is a common pattern, observed in robust classifiers, that (provable) robustness comes at the cost of some accuracy. However, inspecting the found $\varepsilon_{\text{max}}$, we see that the \system agent is much more robust. The $\varepsilon_{\text{max}}$ values in \cref{tab:diffai} have been multiplied by 255, to be on the same scale as discrete intensity changes of pixels in the image. Obtaining values above 1 means each pixel value in the frame can be changed by $\pm 1$ and the network --- provably --- will still select the same action.

\section{Conclusion} \label{sec:conclusion}
We introduced an extension to the DQN training algorithm that enables us to incorporate state-of-the-art defenses such as adversarial or provably robust training into the learning process. The key idea is to split the learning process among two networks, one of which aims to learn a correct Q-function and another that aims to explore the environment robustly with respect to perturbations. We showed that this algorithm clearly outperforms DQNs in terms of both, adversarial and provable robustness.

\message{^^JLASTBODYPAGE \thepage^^J}

\clearpage
\bibliographystyle{ieeetr}
\bibliography{references}

\message{^^JLASTREFERENCESPAGE \thepage^^J}

\clearpage
\appendix

{ 
    \centering
    {\LARGE\bf Supplementary Material for\\Online Robustness Training for Deep Q Learning\par}
  }

\section{Hyperparameters} \label{sec:hyper}
In \cref{tab:hyperparameters} we provide the hyperparameters used for the experiments throughout the paper.
We found the parameters by starting from the parameters in RainbowDQN \cite{rainbowDQN} and manually tweaking them for a fast training DQN. We then used the same parameters including network architecture (plus additional \system-specific parameters) for \system. Further, we found training the algorithm to expensive for automatic hyperparameter tuning (\eg Bayesian Optimization).
\begin{table*}[h]
\centering
\caption{Hyperparameters used in the experiments. $\to$ indicates linear annealing.}
\footnotesize 
\begin{tabu}{lccccc}
  & \textbf{FreeWay}
  & \textbf{Bank Heist}
  & \textbf{Pong}
  & \textbf{Boxing}
  & \textbf{Road Runner}\\ \midrule
Optimizer &  \multicolumn{5}{c}{Adam}  \\ \clinelight{1-}
Adam-$\epsilon$ & \multicolumn{5}{c}{0.00015}\\ \clinelight{1-}
Learning rate $\eta_{Q}$ &  0.0001 &  0.0003 &  0.0001 &  0.0001&  0.0001  \\ \clinelight{1-}
Learning rate $\eta_S$ &  \multicolumn{5}{c}{0.0002}  \\ \clinelight{1-}
Batch-size & \multicolumn{5}{c}{32}\\ \clinelight{1-}
Clip reward to sign & \multicolumn{5}{c}{True} \\ \clinelight{1-}
Double Q-learning & \multicolumn{5}{c}{True}\\ \clinelight{1-}
Game played for & \multicolumn{5}{c}{4000000 frames}\\ \clinelight{1-}
Frame-Stack &  \multicolumn{5}{c}{4}\\ \clinelight{1-}
Discount factor $\gamma$ &  \multicolumn{5}{c}{0.99} \\ \clinelight{1-}
Use Priority Replay &  \multicolumn{5}{c}{True} \\ \clinelight{1-}
Priority Replay $\alpha$ (or \(\omega\)) &  \multicolumn{5}{c}{0.5}  \\ \clinelight{1-}
Target net Sync \(m\) &  \multicolumn{5}{c}{every 2000 frames}  \\ \clinelight{1-}
NoisyNet Explore Constant $\kappa$ & \multicolumn{5}{c}{4} \\ \clinelight{1-}
Frames before learning & \multicolumn{5}{c}{80000} \\ \clinelight{1-}
Size of replay buffer & \multicolumn{5}{c}{200000} \\ \clinelight{1-}
$\epsilon$-greedy exploration & \multicolumn{5}{c}{$1.0 \to 0.0$ over 20000 frames} \\
\end{tabu}
\label{tab:hyperparameters}
\end{table*}

\section{DiffAI Command} \label{sec:diffaicommand}
We use the following invocation to train the advantage part of \system:
\begin{lstlisting}
LinMix(a=Point(),
       b=InSamp(Lin(0, 1, 3500000, 500000),
                w=Lin(0, 1.0/255.0, 3500000, 500000)),
       bw=Lin(0, 0.5,  3500000, 500000))  
\end{lstlisting}
We pass the current number of frames played as the progress counter for \lstinline{Lin} to DiffAI.

\section{Further Evaluation Results} \label{sec:results_details}

\begin{table*}[t]
\tiny
\centering
  \caption{Average  evaluation scores ($\pm$ standard deviation) over 15 games with and without training and test time attacks. \system is defended with Adversarial Training. We note here that Pong and Boxing allow negative scores. Further, the high variance in Bank Heist and Road Runner can explained as these games have multiple levels. Usually the agent only learns to solve the first one. When arriving at the next level (in the next episode). We see that usually the agent obtains a good score in one level equal to roughly twice the mean and then zero (or close to it) in the next, yielding the mean and high standard deviation.}
\vspace{5pt}
\begin{tabu}{@{}llrrrr@{}}

& & \multicolumn{2}{c@{}}{no training attack} & \multicolumn{2}{c@{}}{\emph{TrainingPGD(k=1)}}\\
\cmidrule(lr@{0pt}){3-4}
\cmidrule(lr@{0pt}){5-6}

Game
& TestAttack
& DQN
& \system
& DQN
& \system\\

\midrule

\multirow{3}{*}{Freeway} & none    & $33.00 \pm 0.65$ &  $29.33 \pm 0.98$ & $21.73 \pm 1.70$ & $32.93 \pm 0.70$\\
                         & \emph{TestPGD(k=1)} &  $0.00 \pm 0.00$ &  $28.40 \pm 0.99$ & $23.07 \pm 2.49$ & $32.20 \pm 0.86$\\
                         & \emph{TestPGD(k=4)} &  $0.00 \pm 0.00$ & $27.93 \pm 1.16$ & $22.53 \pm 2.53$ & $32.53 \pm 1.06$\\
                         & \emph{TestPGD(k=50)} & $0.00 \pm 0.00$ & $28.13 \pm 1.77$ & $21.20 \pm 1.52$ & $31.80 \pm 1.15$\\
                         \clinelight{1-}
\multirow{3}{*}{Bank Heist} & none    & $222.00 \pm 239.23$ &  $112.66 \pm 165.37$ & $220.00 \pm 213.47$ & $238.66 \pm 253.03$\\
                            & \emph{TestPGD(k=1)} &  $17.33 \pm 16.67$  &   $97.33 \pm 148.10$ & $138.67 \pm 192.42$ & $199.33 \pm 265.58$\\
                         & \emph{TestPGD(k=4)} &  $3.33 \pm 4.88$ & $121.33 \pm 172.49$ & $45.33 \pm 50.97$ & $190.67 \pm 223.49$ \\
                         & \emph{TestPGD(k=50)} &  $2.67 \pm 4.58$ & $160.00 \pm 197.84$ & $42.00 \pm 111.43$ & $132.00 \pm 174.60$\\
                         \clinelight{1-}
\multirow{3}{*}{Pong} & none    &  $20.20 \pm 2.85$ & $16.73 \pm 2.89$  &  $20.46 \pm 1.06$ & $19.73 \pm 0.96$\\
                      & \emph{TestPGD(k=1)} & $-21.00 \pm 0.00$ & $16.20 \pm 2.86$  & $-5.80\pm 4.16$ & $16.33 \pm 4.32$\\
                         & \emph{TestPGD(k=4)} &  $-20.73 \pm 0.46$ & $15.87 \pm 2.32$ & $-12.87 \pm 4.88$ & $18.13 \pm 2.166$\\
                         & \emph{TestPGD(k=50)} &  $-21.00 \pm 0.00$ & $16.80 \pm 3.52$ & $-13.67 \pm 3.46$ & $18.67 \pm 1.29$\\
                         \clinelight{1-}
\multirow{3}{*}{Boxing} & none    &  $95.87 \pm 8.00$ & $93.27 \pm 6.96$ & $84.80 \pm 10.62$ & $80.67 \pm 13.32$\\
                        & \emph{TestPGD(k=1)} &   $7.40 \pm 10.63$ & $62.87 \pm 28.96$ & $10.60 \pm 12.05$ & $58.73 \pm 17.28$\\
                         & \emph{TestPGD(k=4)} &  $-2.80 \pm 4.44$ & $70.33 \pm 21.70$ & $9.20 \pm 13.15$ & $50.87 \pm 25.30$\\
                         & \emph{TestPGD(k=50)} &  $3.80 \pm 7.14$ & $66.93 \pm 26.49$ & $8.87 \pm 14.03$ & $59.40 \pm 18.90$\\
                         \clinelight{1-}
\multirow{3}{*}{Road Runner} & none    &  $9406.67 \pm 13596.14$ & $11920.00 \pm 16798.99$ & $7066.67 \pm 9685.60$ & $12106.67 \pm 15816.74$\\
                        & \emph{TestPGD(k=1)} &   $0.00 \pm 0.00$ & $11560.00 \pm 12354.44$  & $1313.33 \pm 1472.05$ & $5180.00 \pm 6162.47$\\
                         & \emph{TestPGD(k=4)} &  $0.00 \pm 0.00$ & $9293.33 \pm 12067.04$  & $1266.67 \pm 984.64$ & $5753.33 \pm 7632.25$\\
                         & \emph{TestPGD(k=50)} &  $0.00 \pm 0.00$ & $7660.00 \pm 10213.21$  & $1246.67 \pm 1100.56$ & $5146.67 \pm 7310.55$\\
                       \end{tabu}
\label{tab:score_full}
\end{table*}

\begin{table*}[t]
\tiny
\centering
\caption{Score and size $\varepsilon$ of the largest provable region $\mathbb{B}_{\varepsilon}(s)$ around the state $s$. Both numbers are averaged ($\pm$ standard deviation) over 15 games. Values for $\varepsilon$ are found via binary search. \system uses provably robust training. For a discussion for standard deviation see \cref{tab:score_full}}

  \vspace{5pt}
\begin{tabu}{@{}lrrrr@{}}

  & \multicolumn{2}{c@{}}{Score}
  & \multicolumn{2}{c@{}}{$\varepsilon_{\text{max}}$ }\\
\cmidrule(lr@{0pt}){2-3}
\cmidrule(lr@{0pt}){4-5}

Game
& DQN
& \system
& DQN
& \system\\

\midrule

Freeway     & $33.00 \pm 0.65$    &  $32.53 \pm 0.92 $  & $3.39 \cdot 10^{-7} \pm 3.78 \cdot 10^{-8}$ & $7.95 \cdot 10^{-3} \pm 5.01 \cdot 10^{-5}$\\
Bank Heist  & $222.00 \pm 239.23$   & $154.00 \pm 190.18$  & $5.55 \cdot 10^{-6} \pm 2.49 \cdot 10^{-6}$ & $5.38 \cdot 10^{-3} \pm 1.53 \cdot 10^{-3}$\\
Pong        &  $20.20 \pm 2.85$   &   $5.13 \pm 14.01$  & $6.71 \cdot 10^{-7} \pm 1.49 \cdot 10^{-7}$ & $4.21 \cdot 10^{-3} \pm 2.13 \cdot 10^{-4}$\\
Boxing      &  $95.87 \pm 8.00$   & $90.47 \pm 8.68$   & $1.75 \cdot 10^{-7} \pm 1.93 \cdot 10^{-8}$ & $5.42 \cdot 10^{-3} \pm 5.61 \cdot 10^{-4}$\\
Road Runner &  $9406.67 \pm 13596.14$ & $5166.67 \pm 5714.98$ & $1.30 \cdot 10^{-7} \pm 1.06 \cdot 10^{-7}$ & $4.38 \cdot 10^{-3} \pm 2.43 \cdot 10^{-4}$\\

\end{tabu}
\label{tab:diffai_full}
\end{table*}

\message{^^JLASTPAGE \thepage^^J}

\end{document}